\documentclass[a4paper,fleqn]{cas-dc}

\usepackage[numbers]{natbib}

\def\tsc#1{\csdef{#1}{\textsc{\lowercase{#1}}\xspace}}
\tsc{WGM}
\tsc{QE}
\tsc{EP}
\tsc{PMS}
\tsc{BEC}
\tsc{DE}

\begin{document}
\let\WriteBookmarks\relax
\def\floatpagepagefraction{1}
\def\textpagefraction{.001}
\shorttitle{}
\shortauthors{Md Fazle Rabby et~al.}

\title [mode = title]{Stacked LSTM Based Deep Recurrent Neural Network with Kalman Smoothing for Blood Glucose Prediction}                      



\newcommand{\XH}[1]{{\authnote{\textcolor{blue}{HEI}}{\textcolor{red}{#1}}}}
\newcolumntype{Z}{>{\centering\arraybackslash}m{0.95cm}}

\author[1]{Md Fazle Rabby}[type=editor,
                        auid=000,bioid=1,
                        prefix=,
                        role=,
                        orcid=]
\cormark[1]
\fnmark[1]
\ead{mdfazle.rabby1@louisiana.edu}
\address[1]{The University of Louisiana at Lafayette, Lafayatte, LA 70503}
\author[1]{Yazhou Tu}[style=chinese]
\author[1]{Md Imran Hossen}[style=chinese]
\author%
[2]{Insup Lee.}
\address[2]{University of Pennsylvania, Philadelphia, PA 19104}
\author[1]{Anthony S Maida}[style=chinese]
\author[1]{Xiali Hei}[style=chinese]



\begin{abstract}
Blood glucose (BG) management is crucial for type-1 diabetes patients resulting in the necessity of reliable artificial pancreas or insulin infusion systems. In recent years, deep learning techniques have been utilized for a more accurate BG level prediction system. However, continuous glucose monitoring (CGM) readings are susceptible to sensor errors. As a result, inaccurate CGM readings would affect BG prediction and make it unreliable, even if the most optimal machine learning model is used. In this work, we propose a novel approach to predicting blood glucose level with a stacked Long short-term memory (LSTM) based deep recurrent neural network (RNN) model considering sensor fault. We use the Kalman smoothing technique for the correction of the inaccurate CGM readings due to sensor error. For the OhioT1DM dataset, containing eight weeks' data from six different patients, we achieve an average RMSE of 6.45 and 17.24 mg/dl for 30 minutes and 60 minutes of prediction horizon (PH), respectively. To the best of our knowledge, this is the leading average prediction accuracy for the ohioT1DM dataset. Different physiological information, e.g., Kalman smoothed CGM data, carbohydrates from the meal, bolus insulin, and cumulative step counts in a fixed time interval, are crafted to represent meaningful features used as input to the model. The goal of our approach is to lower the difference between the predicted CGM values and the fingerstick blood glucose readings - the ground truth. Our results indicate that the proposed approach is feasible for more reliable BG forecasting that might improve the performance of the artificial pancreas and insulin infusion system for T1D diabetes management.
\end{abstract}




\maketitle

\section{Introduction}
Diabetes Mellitus is a chronic disorder associated with abnormally high levels of blood glucose because the body is unable to produce enough insulin to meet its needs. $\alpha$-cell and $\beta$-cell in the pancreas are responsible for maintaining the glucose level in blood by secreting insulin and glucagon hormones \cite{national1979classification}. Diabetes can be classified primarily into two categories. Type-1 diabetes is due to $\beta$-cell destruction and would cause absolute insulin deficiency. Type-2 diabetes is due to a progressive insulin secretory defect on the background of insulin resistance \cite{american20172}. In type-1 diabetes, hypoglycemia sets in when blood sugar levels are too low (blood glucose concentration \textless 70 \textit{mg/dl}) \cite{cryer2003hypoglycemia} and hyperglycemia occurs when blood sugar levels are too high for a prolonged period ( blood glucose concentration \textgreater 180 \textit{mg/dl}) \cite{kotagal2015perioperative}, \cite{facchinetti2012online}, \cite{zavitsanou2015silicoclosed}. In the long term, hyperglycemia causes severe complications of heart, blood vessel, eyes, kidneys, and other organs \cite{american2010diagnosis}. Therefore, proper diabetes management is vital for human health.

External insulin treatments are indispensable for T1D diabetes patients to maintain the blood glucose level in a healthy range \cite{vettoretti2017type}. With the naive approach for diabetes management, a patient needs to measure BG concentration several times throughout the day and night using the finger-stick test. Currently, it is the most common self-monitoring approach. Improved techniques such as a combination of an insulin pump for Continuous Subcutaneous Insulin Infusion (CSII) and a device for continuous glucose monitoring (CGM), are required for an effective blood glucose management system, known as sensor-augmented-pump (SAP) therapy \cite{bergenstal2010effectiveness}. CGM device takes glucose measurements with an interval of a particular time-frame. For example, most of the CGM devices take 288 measurements per day with five minutes interval. SAP therapy has been further improved by utilizing control algorithms for dynamic insulin delivery. In terminology, that is known as ``Artificial Pancreas'' (AP), closed-loop control of blood glucose in diabetes \cite{cobelli2011artificial}. Statistical and machine learning techniques with the availability of previous continuous BG records make BG prediction more convenient. This prediction mechanism allows a patient or control algorithm to take the initiative to lower the adverse effect of unintended glycemic events.

Various statistical and machine learning methodologies have been proposed for blood glucose forecasting. The autoregressive integrated moving average (ARIMA) model-based algorithm is an example of a classical statistical method for blood glucose prediction \cite{yang2018arima}. The primary limitation of naive machine learning approaches is that it fundamentally depends on the representation of the input data (e.g., support vector regression \cite{georga2012multivariate} or k-nearest neighbor classifier \cite{karegowda2012cascading}). Daskalaki \textit{et al.}, \cite{bunescu2013blood} proposed a generic physiological model of blood glucose dynamics to generate essential features of a Support Vector Regression (SVR) model that took daily events such as insulin boluses and meals into consideration. The meal model, insulin model, and exercise model were used with the SVR model in work \cite{georga2011glucose}.

Several works have been done for blood glucose prediction using artificial neural networks (ANNs) \cite{perez2010artificial}, \cite{zecchin2012neural}, \cite{pappada2011neural}, \cite{tresp1999neural}. Classically, an ANN has a single hidden layer. However, deep learning models have many hidden layers. These deep models with higher complexity outperform traditional shallow neural networks. These deep learning models with higher complexity can learn the pattern automatically from data \cite{schmidhuber2015deep}, \cite{glorot2010understanding}. Especially for sequential data, recurrent neural networks (RNNs) outperform feed-forward ANNs. However, the drawbacks of using classical RNNs are limitations in its ability to discover, that arises from the vanishing/ exploding gradient problem. This issue has been addressed by Long short-term memory (LSTM) networks \cite{hochreiter1997long} with the addition of memory cell and forget gate to classical RNN. In very recent works,  \cite{li2019convolutional}, \cite{chen2018dilated},  \cite{martinsson2018automatic}, \cite{allam2011recurrent}, convolutional RNN and LSTM are investigated for BG prediction more accurately. Fox \textit{et al.} \cite{fox2018deep} used RNN with Gated Recurrent Unit (GRU) cell and Sun \textit{et al.} \cite{sun2018predicting} utilized Bi-LSTM based RNN for BG forecasting. However, the accuracy achieved with state-of-the-art models for real patient data is not high enough that these approaches might not be applicable in the health domain. For example, the best average RMSE value achieved for the OhioT1DM dataset is 19.04mg/dl so far in the most recent works. Moreover, in these recently proposed methodologies with the RNN model, sensor fault is not taken into consideration. Comparatively, the less accuracy in the BG prediction schemes motivates us for this work to improve the reliability and accuracy of the BG forecasting mechanism.

Early detection and prevention of potential glycemic events such as hypoglycemia and hyperglycemia are one of the primary purposes of the artificial pancreas system. Comparatively, more extended works have been done associated with hypoglycemia detection than the detection of hyperglycemic events. Several statistical methods were studied, such as linear prediction models \cite{cameron2008statistical}, recursive auto-regressive partial least squares models \cite{bayrak2013hypoglycemia} and multi-variable model \cite{turksoy2013hypoglycemia}, for modeling the CGM sensor data for a reliable hypoglycemia early alarm system. In work \cite{daskalaki2013early}, an RNN model and two different autoregressive models were proposed to design a hypoglycemia/hyperglycemia early warning system (EWS). The author developed a time-sensitive ANN-based hypoglycemia prediction system that can predict future hypoglycemic events within a prediction horizon (PH) of 30 minutes \cite{eljil2016predicting}. Additional physiological models, along with ANN, are studied in work \cite{bertachi2018prediction} to predict nocturnal hypoglycemia. However, we see very limited works explicitly for the detection of hyperglycemia \cite{nguyen2014neural}. The authors have shown in this work that the electrocardiographic (ECG) signals can be employed with the ANN model to detect hyperglycemic events practically and non-invasively. 

CGM sensor reading is a crucial factor in BG prediction as a slight error in CGM sensor reading might result in the wrong prediction. However, sensor fault is very common in the CGM system. That is why most of the clinical BG data sets are prone to have errors. Several factors are responsible for that kind of fault, such as the decay of sensor sensitivity, pressure-induced sensor attenuation (PISA) \cite{facchinetti2016modeling}, and interruption in signal transmission, etc. Furthermore, bias and latency might be present in a CGM reading \cite{mahmoudi2016application}. Another issue is that the reading of sensors from the same manufacturer might be different due to manufacturing variability. As a consequence, the predicted BG value and the estimation for insulin might be erroneous due to such faulty BG reading. Consequently, it might decrease the efficacy of the diabetes management system. So, to propose more reliable BG forecasting methodologies, sensor fault should be taken into consideration. However, synthesized data sets are exceptions of this case as this is out of the scope of the sensor fault. 

Moreover, there might be another issue responsible for inaccurate BG prediction. The CGM sensor glucose is measured from interstitial fluid samples instead of the blood sample \cite{medtronicdiabetes}. However, there is a discrepancy in time and magnitude between BG and interstitial glucose (IG) \cite{kulcu2003physiological}. In the BG forecasting system, CGM sensor readings are used as one of the primary inputs to predict the future BG level. As IG and BG values are different, the BG prediction with IG as input is challenging. We find in our experiment that the BG prediction made with the Kalman smoothed CGM reading is closer to actual BG reading measured with fingerstick than the prediction made with unprocessed CGM reading. In this study, we assume the fingerstick BG reading as the ground truth.

\begin{figure}[htp]
    \centering
     \includegraphics[scale=.40]{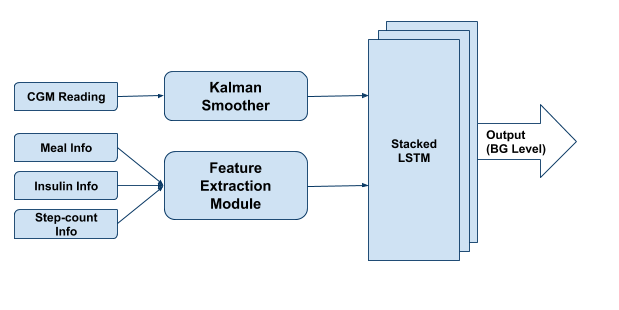}
     \vspace{-2mm}
     \caption{Architecture of the proposed BG prediction system} 
     \label{fig:proposed_archi}
\end{figure}

In this work, we propose a deep learning approach for blood glucose prediction using a multi-layered LSTM based recurrent neural network model. We use OhioT1DM dataset \cite{marling2018ohiot1dm}, containing eight weeks' data for each of six people with type-1 diabetes, for training and evaluation of our proposed model. We utilize several features instead of only CGM measurement. This is because the dynamics of the glucoregulatory system depend on several factors such as carbohydrate intake from meals, the amount of bolus dose or infused insulin, exercise, and physical activeness. Using only the CGM measurement as input might not be enough for learning those complex dynamics. Therefore, we investigate several combinations of different physiological information from the OhioT1DM dataset and choose the most optimal combination of those for our proposed model. We find that step count information from a fitness band can be a useful feature. It turns out that the combination of CGM reading, step count, carbohydrate intake from the meal, and bolus dose, is the most optimal feature set as the input of the model. Note that we use preprocessed CGM data with Kalman smoothing (KS) \cite{staal2018kalman} instead of using raw CGM data to mitigate the effect of sensor fault on BG prediction. The overall architecture of our proposed BG prediction system has been illustrated in figure 1. 

Our main contribution can be summarized as follows:
\begin{itemize}
    \item We propose a novel approach using a stacked LSTM based deep recurrent neural network architecture for BG prediction.
    \item We introduce a more reliable BG prediction system by utilizing Kalman smoothing for mitigating the sensor fault in CGM reading. We demonstrated that the prediction made with this approach is closer to the actual BG level (fingerstick BG reading) than the conventional prediction method. 
    \item Our study reveals that a person's step count information from the fitness band can be effectively utilized for improving BG prediction accuracy. 
\end{itemize}

Overall, our proposed methodologies provide more accurate and reliable forecasting accuracy in terms of RMSE.


\section{Data Processing} \label{data-processing}
\subsection{Dataset} \label{dataset}
We use the OhioT1DM \cite{marling2018ohiot1dm} dataset to train and test our model. This dataset includes eight weeks of data for each of six type-1 diabetes patients. The number of male and female patients was two and four, respectively. For data collection, Medtronic 530G insulin pumps and Medtronic Enlite CGM sensors were used throughout the data collection period. Each of the patients reported daily events data via a smartphone app and a fitness band. The dataset includes CGM blood glucose level every 5 minutes - 288 samples per day, blood glucose levels with fingerstick (self-monitoring), insulin doses, in the form of bolus and basal, self-reported meal time with estimated carbohydrate intake, exercise time, sleep, work, stress, and illness. The data set also includes 5-minute aggregations of step count, heart rate, galvanic skin response (GSR), skin, and air temperature. In this study, we experiment with each of these attributes to find out the optimal attribute set for the BG prediction model. Table \ref{ohiot1dm-dataset} shows the number of training and test examples for each patient.

\begin{table}[htbp]
\renewcommand{\arraystretch}{1.3}
\caption{Gender, number of training and test examples per patient} 
\label{ohiot1dm-dataset} \vspace{-1mm}
\small
\centering
\begin{tabular}{ >{\centering\arraybackslash}m{1.3cm}  | >{\centering\arraybackslash}m{1.7cm} | >{\centering\arraybackslash}m{1.7cm} | >{\centering\arraybackslash}m{1.7cm} }
    \hline
    \bf{Patient ID} & \bf{Gender} &  \bf{Training Examples} & \bf{Test Examples}\\
    \hline \hline
    \# 559    & female    &  10796  & 2514 \\
    \hline
    \# 563    & male   & 12124  & 2570 \\
    \hline
    \# 570    & male    & 10982	& 2745 \\
    \hline
    \# 575	& female   & 11866	    & 2590 \\
    \hline
    \# 588	& female   & 12640   & 2791 \\
    \hline
    \# 591	& female   & 10847	& 2760 \\
    \hline \hline
\end{tabular} \vspace{-4mm}
\end{table} \vspace{1mm}

Initially, we determine the accuracy of the model trained with only the single feature - BG values. We assume it as base accuracy. We trial with each of the attributes only with CGM values separately and compare the accuracy with base accuracy. If the new accuracy gets better than the base accuracy, we consider that attribute as prospective. Otherwise, we exclude it. From our experiment, it turns out that CGM values, carbohydrate intake from the meal, insulin dose as a bolus and 5-minutes aggregation of step count from the fitness band, have a positive effect on the accuracy of the model. It is worth mentioning that some attributes such as insulin from basal, sleep, heart rate, GSR, and skin temperature have no effect or sometimes adverse effect on the accuracy improvement of our proposed model. 

\subsection{Feature Extraction from Physiological Information} \label{feature-extraction}
We select CGM values, meal info, insulin dose, and step count info, as the final feature set. These four features constitute the 4-channel inputs for the proposed model.

\subsubsection{Carbohydrate Information} \label{carbs-info}
In the case of meal information, we consider the fact that blood sugar starts to rise after 15 minutes of having the meal and reaches its peak after 1 hour \cite{kraegen1981timing}. We name this phase as the \textit{Increasing Phase}. Then the carbohydrate level in blood starts to decrease. That is the \textit{Decreasing Phase}. At any time-index $t_s$, we calculate the amount of carbohydrate that is effective at that moment in the blood after having a meal. The calculated amount of carbohydrates is treated as the input for the time-index $t_s$ to the model. Every time the subject encounters a meal, we need to keep track and update the time-index $t_{meal}$ and the amount of carbohydrate $C_{meal}$ from the meal. We ignore the first 15 minutes (3 samples of data-point with 5 minutes of an interval) right after having a meal since it takes 15 minutes to have the effect of the meal on the blood glucose level. Thus, the equation for the effective carbohydrates $C_\mathit{eff}$ in blood for any time-index $t_s$ within the first 60 minutes of time interval after having the meal, is as follows:

\vspace{-1mm}
\begin{equation}\label{carbs_increasing_1}
    \begin{small} 
    \begin{array}{l}
        C_{\mathit{eff}}(t_s) = \{(t_s-t_{\mathit{meal}})\beta_{\mathit{inc}}\}C_{\mathit{meal}}
    \end{array} 
    \end{small}
\vspace{-1mm}
\end{equation}

Here $\beta_{inc}$ is \textit{carbs increasing factor}. We use $\beta_{inc} = 0.111$, which implies that the carbohydrate level in the blood reaches its maximum (100\% of the amount of carbohydrate taken from the meal) by increasing with the rate of 11.1\% for every increment of time-index (5 minutes interval). Note that, $t_{meal} >= t_s$. At the 60th minute after having the meal, $C_\mathit{eff}$ attains the maximum value by increasing with a rate of 11.1\% per time-index. At that moment, the value of $(t_s-t_{meal})\beta_{inc}$ has been approximately less than or equal to 1.00. Note that we do not consider the first 15 minutes (3 time-indexes) right after having the meal in the above equation; instead, we consider nine time-indexes out of twelve time-indexes withing 60-minutes interval so that the value of $C_\mathit{eff}$ doesn't exceed 100\% of $C_\mathit{meal}$. When $C_\mathit{eff}$ reaches its maximum, then the decreasing phase begins. We calculate the amount of carbohydrate effective or appeared in the blood as glucose at any particular time-index $t_s$ within the decreasing phase is as follows.

\vspace{-1mm}
\begin{equation}\label{carbs_decreasing}
    \begin{small} 
    \begin{array}{l}
        C_\mathit{eff}(t_s) = C_{\mathit{meal}}\{1 - (t_s-t_{\mathit{meal}})\beta_{\mathit{dec}} \}
    \end{array} 
    \end{small}
\vspace{-1mm}
\end{equation}

Here, $\beta_{dec}$ is the \textit{carbs decreasing factor}. we set the value of $\beta_{dec}$ as 0.028 according to the assumption that the duration of the decreasing phase is around 3 hours. This means that after 3 hours (the number of time-index is 36), $C_\mathit{eff}$ would be approximately near to 0.00. We update $C_\mathit{eff}$ in every steps where, $C_\mathit{eff} = max(0, C_\mathit{eff})$. Thus, we ignore any negative values for $C_\mathit{eff}$.

\subsubsection{Insulin Information} \label{insulin-info}
We use crafted insulin information as one of the four inputs to the proposed RNN model. We consider only bolus information for further crafting. In this study, we find that basal insulin has no considerable effect on blood glucose prediction accuracy. There are two pathways of insulin absorption \cite{wilinska2004insulin}, slow and fast. Approximately 67\% of delivered insulin passed through the slow channel with an average absorption rate of 0.011 $min^{-1}$, whereas 33\% of insulin passed through the fast one with an absorption rate of 0.021 $min^{-1}$. Therefore, in this experiment, we use the weighted average of these two absorption rates, which is 0.014 $min^{-1}$. For every five minutes interval, we denote it with $R_\mathit{insulin} = 5*0.014$. We calculate the effective insulin on the body $I_\mathit{eff}$ at any particular time-index $t_s$ with the equation as follows: 

\vspace{-1mm}
\begin{equation}\label{insulin_decreasing}
    \begin{small} 
    \begin{array}{l}
        I_\mathit{eff}(t_s) = I_\mathit{bolus}\ - (t_s-t_\mathit{bolus})R_\mathit{insulin}
    \end{array} 
    \end{small}
\vspace{-1mm}
\end{equation}

Here, $I_\mathit{bolus}$ is the amount of insulin delivered in the form of bolus to the patient, and $t_\mathit{bolus}$ is the time-index when the most recent insulin delivered to the patient. We update $I_\mathit{eff}$ in every steps, where $I_\mathit{eff} = max(0, I_\mathit{eff})$. This calculated effective insulin in the body $I_\mathit{eff}$ is used as one of the inputs for the RNN model. In the above equation, we do not consider insulin absorption delay. 

\subsubsection{Step Count Information} \label{step-count-info}
We compute a weighted average of the number of steps taken by the patient at time $t_s$. To calculate $S_{avg}$, we consider the previous 50 minutes (10 readings with 5 minutes of the interval) where the steps count for most recent time-index has a more significant weight. Note that the weight decreases gradually with time-index. 

\vspace{-1mm}
\begin{equation}\label{insulin_decreasing}
        S_{avg}(t_s) =  \frac{1}{n} \sum_{i=0}^{n-1} (n-i) \times steps(t_s - i)
\vspace{-1mm}
\end{equation}

In the above equation, $n=10$ as we only consider the previous 50 minutes or 10 data points. The computed $ S_{avg}$ is treated as one of the inputs.

\begin{figure*}[pos=htp,width=\textwidth,align=\centering]
    \centering
     \includegraphics[scale=.45]{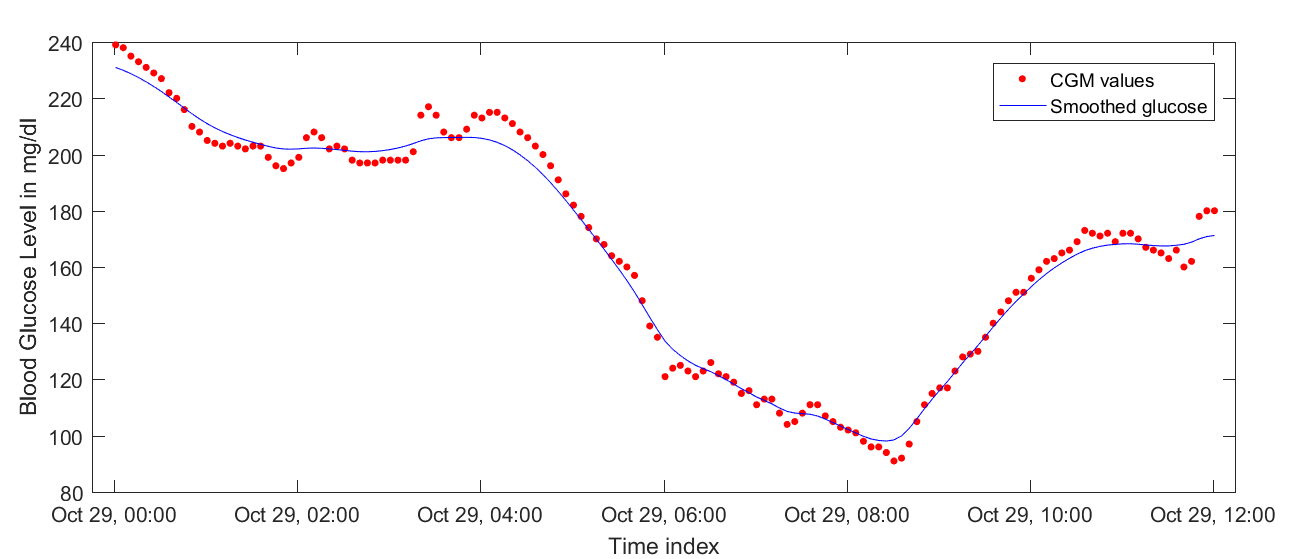}
     \vspace{-1mm}
     \caption{Kalman smoothed CGM values vs raw CGM readings for 12 hours time window from the patient \#563. Here red dots are raw CGM readings, where the blue line is the smoothed CGM values} 
     \label{fig:Kalman_Smoothing}
\end{figure*}

\subsubsection{Glucose Level Information from CGM readings}
Either unprocessed or Kalman smoothed CGM readings are considered as an input channel out of the four inputs.

\subsection{Kalman Smoothing for preprocessing} \label{kalman-smooting}
In this section, we discuss Kalman filtering and Kalman smoothing briefly. The KS method outputs an interpolated time series of glucose estimates with mean and variance. It can automatically correct errors in the CGM readings where the estimated variance can be utilized for determining at which times the data are reliable. In our study, KS has been used as a pre-processing technique for sensor fault correction in the CGM reading. We use a modified implementation of KS for the OhioT1DM dataset, from the work \cite{staal2018kalman}.

The Kalman filter is a technique of estimating the current state of a dynamical system from the previous observations. In Kalman filtering, records of data are used for the calculation of the estimates. Thus, the Kalman filter is appropriate for real-time data processing. It is a forward algorithm where each step is computed analytically. The model and observation can be written as:

\vspace{-3mm}
\begin{equation}\label{eq_1_initialization}
    \begin{small} 
    \begin{array}{l}
        x_{k+1} = \phi_{k}x_{k} + B_{k}u_{k} + w_k
    \end{array} 
    \end{small}
\vspace{-1mm}
\end{equation}

\vspace{-4mm}
\begin{equation}\label{eq_2_initialization}
    \begin{small} 
    \begin{array}{l}
        y_{k} = H_{k}x_{k} + v_k
    \end{array} 
    \end{small}
\vspace{-1mm}
\end{equation}

Here, $x$, $u$, and $y$ are the system internal state, input to the system, and measured output respectively. Whereas $v$ is the process noise, and $w$ is the measurement noise. These noise processes are assumed to be zero-mean Gaussian. $\phi$ is the transition matrix, and $H$ is the measurement matrix.

In the phase of time update, the Kalman filter computes the \textit{priori estimates}, a state estimate $\bar x$ and state covariance matrix $\bar P$ \cite{rauch1965maximum}.

\vspace{-3mm}
\begin{equation}\label{eq_3_priori}
    \begin{small} 
    \begin{array}{l}
        \bar x_k =  \phi_{k-1}\hat x_{k-1} + B_{k-1}u_{k-1}
    \end{array} 
    \end{small}
\vspace{-1mm}
\end{equation}

\vspace{-4mm}
\begin{equation}\label{eq_4_priori}
    \begin{small} 
    \begin{array}{l}
        \bar P_k =  \phi_{k-1}\hat P_{k-1}\phi^T_{k-1} + Q_{k-1}
    \end{array} 
    \end{small}
\vspace{-1mm}
\end{equation}

Then the phase \textit{measure update} is performed, where the \textit{posteriori estimate} $\hat x$ and $\hat P$ are calculated. The equations are as follows \cite{rauch1965maximum}: 

\vspace{-3mm}
\begin{equation}\label{eq_5_posteriori}
    \begin{small} 
    \begin{array}{l}
        K_k =  \bar P_k H^T_k(H_k\bar P_k H^T_k + R_k)^{-1}
    \end{array} 
    \end{small}
\vspace{-1mm}
\end{equation}

\vspace{-4mm}
\begin{equation}\label{eq_6_posteriori}
    \begin{small} 
    \begin{array}{l}
        \hat x_k =  K_k(y_k - H_k \bar x_k)
    \end{array} 
    \end{small}
\vspace{-1mm}
\end{equation}

\vspace{-4mm}
\begin{equation}\label{eq_7_posteriori}
    \begin{small} 
    \begin{array}{l}
        \hat P_k =  (I - K_k H_k)\bar P_k
    \end{array} 
    \end{small}
\vspace{-1mm}
\end{equation}

Kalman smoothing can be applied to get better estimates than Kalman filtering. However, it is required to have the whole dataset available at the time of performing Kalman smoothing. In our experiment, that is true. The Rauch-Tung-Striebel (RTS) algorithm \cite{rauch1965maximum} utilizes previous as well as the following data at the time $k$ to generate the estimate. In RTS, there is one forward pass through the available data applying the Kalman filter to generate the priori, posteriori, and covariance matrices. These generated estimates and covariance are then treated as input to a subsequent backward pass. In this phase, RTS calculates the smoothed estimate $\hat x^s_k$ and $\hat P^s_k$.

\vspace{-3mm}
\begin{equation}\label{eq_8_smoothing}
    \begin{small} 
    \begin{array}{l}
        C_k =  \hat P_k \phi_k \bar P^{-1}_{k+1}
    \end{array} 
    \end{small}
\vspace{-1mm}
\end{equation}

\vspace{-4mm}
\begin{equation}\label{eq_9_smoothing}
    \begin{small} 
    \begin{array}{l}
        \hat x^s_k = \hat x_k + C_k(\hat x^s_{k+1} - \bar x_{k+1})
    \end{array} 
    \end{small}
\vspace{-1mm}
\end{equation}

\vspace{-4mm}
\begin{equation}\label{eq_10_smoothing}
    \begin{small} 
    \begin{array}{l}
        \hat P^s_k = \hat P_k + C_k(\hat P^s_{k+1} - \bar P_{k+1})C^T_k
    \end{array} 
    \end{small}
\vspace{-1mm}
\end{equation}

We apply Kalman smoothing on CGM readings from the OhioT1DM dataset to get smoothed CGM values. Original CGM readings (represented by red dots) and smoothed CGM values (represented by the blue line) are shown in Fig. \ref{fig:Kalman_Smoothing}. It is noticeable that, after applying Kalman smoothing, there are less abrupt changes or fluctuations in smoothed CGM values.

\section{Modeling}
Our model uses a neural network to learn the prediction. A recurrent neural network (RNN) is a feed-forward neural network that can model sequential data. It utilizes weight sharing between each element in the sequence over time. There are diverse variants of RNN. In the basic RNN variant termed as Vanilla RNN, the transition function is a linear transformation of the hidden state vector $h$ and the input vector $x$, followed by an activation function for non-linearity. 

\vspace{-3mm}
\begin{equation}\label{eq_modeling_ht}
    \begin{small} 
    \begin{array}{l}
        h_t = \tanh{(W[h_{t-1} , x_t ] + b)}
    \end{array} 
    \end{small}
\vspace{-1mm}
\end{equation}

Where $W$ is weight matrix, $b$ is a bias vector, and $\tanh$ is the activation function. Classical RNN has the form of a chain of repeating modules of neural networks with straightforward architecture. Theoretically, RNN can learn long term dependency. However, practically it suffers from vanishing gradient problem and exploding gradient problem as a result of long term dependency. This dependency makes RNN less useful and more challenging to train \cite{hochreiter1998vanishing}. Hochreiter and Schmidhube \cite{hochreiter1997long} proposed Long short-term memory (LSTM) for addressing this issue. The LSTM network mitigates this long-term dependency problem to some extent by utilizing the concept of memory, the gate structure, and constant error carousel. In the case of our study, LSTM is more suitable to model blood glucose levels as there are dependencies upon immediate previous entries in sequential diabetes patient data. Consequently, we prefer the LSTM based RNN model over the other model architectures to make the BG prediction more rigorous.


In our work, we build an LSTM network containing 128 element hidden vector. Each cell at step $t$ includes a forget gate $f_t$, an input gate $i_t$, a control gate $\widetilde{C_t}$, an output gate $o_t$, and an internal cell memory $C_t$. 
 
The first gate is the forget gate $f_t$ that determines which information can be carried to the cell from the output of the previous LSTM cell. It is a single layer neural network, that can be represented as: 

\vspace{-3mm}
\begin{equation}\label{eq:forget_gate}
    \begin{small} 
    \begin{array}{l}
       f_t = \sigma(W_f[h_{t-1} , x_t ] + b_f) 
    \end{array} 
    \end{small}
\vspace{-1mm}
\end{equation}
 
 
The input gate $i_t$ controls how much the new memory should influence the old memory. It can be written as: 

\vspace{-3mm}
\begin{equation}\label{eq:input_gate_gate}
    \begin{small} 
    \begin{array}{l}
       i_t = \sigma(W_i[h_{t-1} , x_t ] + b_i)
    \end{array} 
    \end{small}
\vspace{-1mm}
\end{equation}

The control gate $\widetilde{C_t}$ generates new memory and updates cell state from $C_{t-1}$ and $\widetilde{C_t}$. Here $\odot$ represents element-wise multiplication.

\vspace{-3mm}
\begin{equation}\label{eq:control_gate_1}
    \begin{small} 
    \begin{array}{l}
       \widetilde{C_t} = \tanh{(W_C[h_{t-1}, x_t ] + b_C)}
    \end{array} 
    \end{small}
\vspace{-1mm}
\end{equation}

\vspace{-4mm}
\begin{equation}\label{eq:control_gate_2}
    \begin{small} 
    \begin{array}{l}
       C_t = f_t \odot C_{t-1} + i_t \odot \widetilde{C_t}
    \end{array} 
    \end{small}
\vspace{-1mm}
\end{equation}

The output gate is responsible for modulating the output to obtain $h_{t-1}$. It can be expressed with the equation below:

\vspace{-3mm}
\begin{equation}\label{eq:output_gate_1}
    \begin{small} 
    \begin{array}{l}
       o_t = \sigma(W_o[h_{t-1} , x_t ] + b_o)
    \end{array} 
    \end{small}
\vspace{-1mm}
\end{equation}

\vspace{-4mm}
\begin{equation}\label{eq:output_gate_2}
    \begin{small} 
    \begin{array}{l}
       h_t = \tanh{(C_t)} \odot o_t 
    \end{array} 
    \end{small}
\vspace{-1mm}
\end{equation}

In the above equations, $W_s$ is the corresponding weight matrix, $\sigma$, and $\tanh$ are sigmoid and hyperbolic tangent activation function, respectively.

We add a dropout layer after the LSTM layer. Dropout is intended to reduce overfitting and improve the generalization of the model \cite{srivastava2014dropout}. The last layer of the LSTM outputs a vector $h_i$, which is fed as the input of a fully connected multi-layer network. This network consists of three layers, including two hidden layers and one output layer. These dense layers contain 512 neurons, 128 neurons, and a single neuron, respectively, with an activation function each. We choose the rectified linear unit (ReLU) activation function for the first two dense layers and the exponential activation function for the output layer. The final output of the network can be represented as follows:

\vspace{-3mm}
\begin{equation}\label{eq:univariate_gaussian}
 \begin{small}
 \begin{array}{l}
    [\mu ,\sigma^2] = activation(W_ih_t + b_i)
 \end{array}
 \end{small}
 \vspace{-1mm}
\end{equation}

Here $\mu$ is the mean, and $\sigma^2$ is the variance. We can evaluate the confidence of the model with these values. The input of the model is a multi-dimensional sequence of preprocessed BG level from CGM reading, carbohydrates amount from the meal, carbohydrates amount from the bolus, and step count related data. The output of the model is a prediction regarding BG level withing a prediction horizon. We experiment with the prediction horizon of 30 and 60 minutes. 

Our proposed model adopts the negative log-likelihood (NLL) loss function as follows:

\vspace{-2mm}
\begin{equation}\label{eq:NLL}
    \begin{small}
    \begin{array}{l}
        L = \frac{1}{k} \sum_{i=0}^{k} -\log(f(y_i | \mu_i, \sigma_i^2 ))
     \end{array}
     \end{small}
     \vspace{-0mm}
\end{equation}

In the above equation, $y_i$ is the target BG value, and $\mu_i$, $\sigma_i$ are the mean and SD respectively for the output of the model for corresponding input $x_i$. The function $f$ represent the difference between target and output BG value. We use Adaptive Moment Estimation (Adam) as the optimizer. 

\begin{figure}[htp]
    \centering
     \includegraphics[scale=.33]{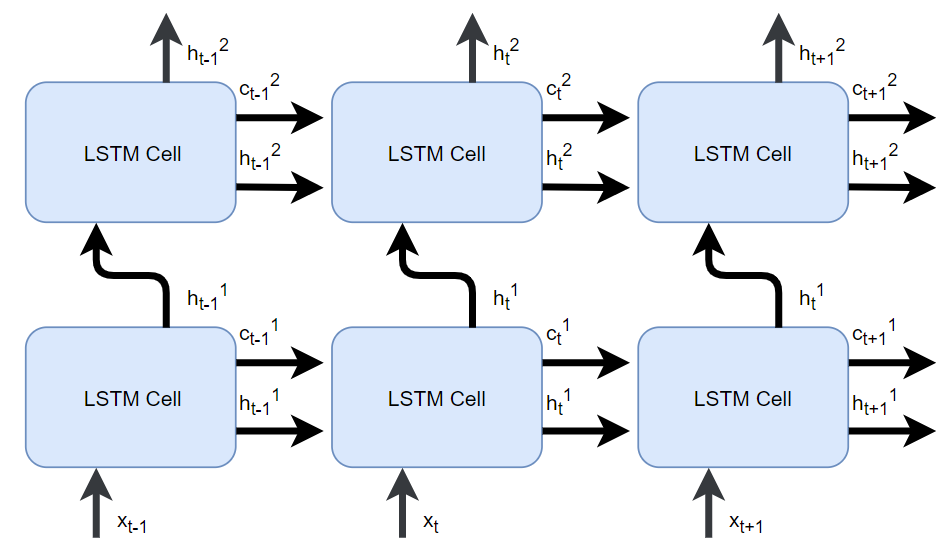}
     \vspace{-2mm}
     \caption{Two Layered Stacked LSTM network} 
     \label{fig:Stacked}
\end{figure}

We study the correlation between BG prediction accuracy and depth of the model architecture. We see in previous work that the deep recurrent neural networks provide empirical superiority over shallow networks \cite{graves2013speech}. The shallow network cannot precisely model the information with a temporal hierarchy \cite{hermans2013training}. However, the concept of depth in an RNN is not the same as it is in feedforward neural networks \cite{pascanu2013construct}. To make the RNN model deeper, we employ the stacking technique. Our proposed model consists of two LSTM layers. The first LSTM layer provides a sequence output that is fed as one input to the LSTM layer above. Both LSTM layers have the same internal architecture described earlier. Fig. \ref{fig:Stacked} illustrates the architecture of two Layered Stacked LSTM. 

We also experimented with the GRU cell instead of the LSTM cell. However, in this application, the network with the LSTM cell outperforms the network with GRU cells. A comparison between the performance of these two types of RNN is shown in section \ref{result}. Thus, we choose the LSTM cell for our final RNN model.

\section{Experimental Setup}
We conduct extensive experiments on data and our proposed model to tune hyperparameters and determine the optimal setup. We measure the efficacy of our proposed method to predict BG values for two different setups. In the first setup, we use raw CGM readings, whereas, in the second setup, we use Kalman smoothed CGM values for sensor error correction.

\subsection{Hyperparameters}
We experiment with two different RNN cell types - LSTM and GRU cells, respectively. The dynamics of the glucoregulatory system are considered as nonlinear. Hence, learning blood glucose dynamics is a complicated task where consideration of previous data is very crucial for effective prediction mechanisms. As a consequence, we skip experimenting with Vanilla RNN cells because of its long term dependency problem \cite{bengio1994learning}. Our experimental result demonstrates that the LSTM network achieves better prediction accuracy than a network with GRU cells. We train our model with the six patients' training data from the OhioT1DM dataset. We experiment with 30, 60, 120, and 240 minutes of data history as the input for the proposed model. Those contain 6, 12, 24, and 48 training examples respectively as CGM readings are taken with every five minutes of interval. We also investigate the effect of the number of LSTM state hidden units and fully connected layers in the network. Different combinations of LSTM state size-of 64, 128, and 248 with two fully connected layers are tested. We find that the network model with 128 LSTM states with two fully connected layers outperforms other configurations. A learning rate of $10^{-3}$ was used for training with a maximum number of 6000 epochs. However, an early-stopping patience threshold of 128 epochs is employed for better convergence.

\subsection{Training, Testing, and Validation}
We use the T1DM dataset for training, validation, and testing purposes. This dataset contains 12 files for six data contributors. For each person, there are two files for training and testing data, respectively. We split the data in the training file into the training and validation dataset. The entire dataset in the testing file is used for testing purposes for each subject. We partition the training data file such that 80\% of the data is used for training, and the rest of the data is used for validation. The purpose of the validation data is to provide an evaluation of the model after hyperparameter tuning.

In the training phase, the prediction from the model is used to determine the subsequent prediction curve at each epoch. It is possible to evaluate the model at a certain point in time if there are at least 24 prior data points are available (prior data points for 2 hours or 120 minutes). This threshold is to assure that there are enough data points available for the model to make a feasible prediction. For the training process, we set the batch size to 128. However, we try with larger batch sizes than 128 and found that it makes the prediction accuracy worse in terms of RMSE. Finally, we train the two instances of the model separately with raw CGM data, and Kalman smoothed CGM data, respectively, along with other features including carbohydrates from the meal, bolus insulin, and cumulative step counts in a fixed time interval. In this study, we consider the previous 50 minutes of history for calculating the cumulative step numbers.

\section{Evaluation Criteria and Results} \label{result}
In this section, we discuss the BG prediction result of our proposed deep RNN model for the T1DM-testing dataset. The performance of the model is compared based on the accuracy over the 30 and 60 minutes prediction horizon. The hyperparameters of the model are tuned extensively for optimal results.

\subsection{Evaluation Criteria}
Initially, we investigate the usefulness of preprocessing the CGM values. We use Kalman smoothing for processing the CGM readings. We compare the deviation of raw CGM readings and preprocessed CGM values from the ground truth reference value. In our analysis, we assume the BG level measured by fingerstick as ground-truth. The glucose readings from the sensor sampling in interstitial fluid are substantially different from blood glucose values measured at the same time \cite{kulcu2003physiological}. As a result, CGM manufacturers suggested that patients should use capillary blood glucose measurements before any treatment decisions. Moreover, the self-measurement of blood glucose (SMBG), has been used as a reference for different CGM systems accuracy comparison \cite{luijf2013accuracy}. Fingerstick testing is one of the most convenient SMBG methods. These are the principal reasons behind our assumption of choosing the fingerstick reading as the ground truth for comparison between raw CGM value and preprocessed CGM values. We used Mean Absolute Error (MAE) to calculate the difference between CGM values (raw CGM, preprocessed CGM) and fingerstick reference BG values at a particular time.

\vspace{-2mm}
\begin{equation}\label{eq:MAE}
    \begin{small}
    \begin{array}{l}
        \mathit{MAE} =  \frac{1}{N} \sum_{i=1}^{N} \mid y_i - x_i\mid
     \end{array}
     \end{small}
     \vspace{-0mm}
\end{equation}
Here, $y$ denotes the CGM values (either raw CGM values or smoothed CGM values), and $x$ denotes the fingerstick reference BG values.

\begin{table}[htbp]
\renewcommand{\arraystretch}{1.3}
\caption{Comparison of the accuracy ($\mathit{MAE}$) of raw CGM values and Kalman Smoothed CGM values with respect to fingerstick BG readings} 
\label{table_RAW_KAL_FING} \vspace{-1mm}
\small
\centering
\begin{tabular}{ >{\centering\arraybackslash}m{1.3cm}  | >{\centering\arraybackslash}m{1.7cm} | >{\centering\arraybackslash}m{1.7cm} | >{\centering\arraybackslash}m{1.7cm} }
    \hline
    \bf{Patient ID} & \bf{\# of Fingerstick} &  \bf{Raw CGM} & \bf{Smoothed CGM}\\
    \hline \hline
    \# 559    & 53    &  21.7  & 19.8 \\
    \hline
    \# 563    & 196   & 16.0  & 14.7 \\
    \hline
    \# 570    & 99    & 10.3	& 9.5 \\
    \hline
    \# 575	& 117   & 9.7	    & 10.5 \\
    \hline
    \# 588	& 391   & 20.8    & 18.4 \\
    \hline
    \# 591	& 197   & 18.7	& 17.4 \\
    \hline \hline
     \textbf{Mean MAE} &  \textbf{N/A}    & \textbf{16.2}	& \textbf{15.1} \\
    \hline
\end{tabular} \vspace{-4mm}
\end{table} \vspace{1mm}

We find that preprocessed CGM values are much closer to fingerstick BG reading than raw CGM values. Table \ref{table_RAW_KAL_FING} illustrates the accuracy comparison, in terms of MAE, between raw CGM and Kalman smoothed CGM values with respect to fingerstick BG readings. Table \ref{table_RAW_KAL_FING} shows that both of the CGM readings (measured from interstitial fluid) and pre-processed CGM values differ from finger stick reading (measure from blood glucose). However, most importantly, processed CGM values have a lower error than raw CGM values. Thus, we conclude that the model trained with Kalman smoothed CGM values along with other features, is more effective in forecasting the BG level. 

However, the performance of a model can be evaluated with different criteria. Among those, the root-mean-square error (RMSE) between the reference CGM values and predicted BG level, is one of the most widely adopted methods to assess the BG prediction accuracy. Thus, we evaluate the performance of our model in terms of RMSE in this paper. 
\vspace{0mm}
\begin{equation}\label{eq11}
 \begin{small}
 \begin{array}{l}
    \mathit{RMSE} = \sqrt{ \frac{1}{N} \sum_{i=1}^{N}(\hat{y}(i|i-PH)^2 - y(i))^2}
 \end{array}
 \end{small}
 \vspace{0mm}
\end{equation}
where ${\hat{y}(i|i - PH)}$ denotes the model's prediction results provided the previous data and $y$ denotes the reference CGM reading, $N$ is the number of data points.  

\subsection{Results for the dataset with unprocessed/raw CGM readings}
In this section, we discuss the BG forecasting accuracy for the proposed stacked RNN model for the OhioT1DM dataset, where CGM readings are raw/unprocessed. Initially, we investigate the effect of the depth of LSTM layers of the network on the prediction accuracy of the model. In our experiment, we consider the prediction horizon of 30 and 60 minutes. The experimental results for models with single LSTM layers and stacked LSTM layers are summarized in table \ref{table_raw_sigle_vs_stacked}. Results from the table demonstrate that the RNN model with stacked LSTM layered architecture performs better than RNN with a single LSTM layer for all of the cases. 

\begin{table}[ht]
\renewcommand{\arraystretch}{1.3}
\caption{The table represents the prediction comparison of proposed models over 30 and 60 minutes of the prediction horizon. Here, the models are trained with the raw CGM readings along with the other three features mentioned before. The first one is the RNN model having a single LSTM layer, whereas the second one is stacked LSTM based deep RNN model.} 

\label{table_raw_sigle_vs_stacked}
\centering
\small
\begin{tabular}{ >{\centering\arraybackslash}m{1.4cm}  | >{\centering\arraybackslash}m{1.2cm} | >{\centering\arraybackslash}m{1.2cm} | >{\centering\arraybackslash}m{1.2cm} | >{\centering\arraybackslash}m{1.2cm} }
    \hline 
    \multirow{3}{*}{ \bf{Patient ID} } &
      \multicolumn{2}{c}{\bf{PH=30 min} } &
      \multicolumn{2}{c}{\bf{PH=60 min} }\\
    \cline{2-5}
    & \bf{Single LSTM}  &   \bf{Stacked LSTM}     &   \bf{Single LSTM}     &   \bf{Stacked LSTM}   \\
    \hline \hline
     \# 559     & 18.03   &     \textbf{17.85}   &   31.89   &   \textbf{31.55}   \\
    \hline
     \# 563     & 19.20   &     \textbf{18.65}   &   31.01   &   \textbf{30.42}   \\
    \hline
     \# 570     & 16.63   &     \textbf{15.94}	&   26.28   &   \textbf{25.74}   \\
    \hline
     \# 575	    & 21.12   &     \textbf{20.93}	&   32.90   &   \textbf{31.97}   \\
    \hline
     \# 588	    & 18.07   &     \textbf{17.71}   &   31.11    &   \textbf{30.45}   \\
    \hline
     \# 591	    & 20.71   &     \textbf{20.35}	&   32.06   &   \textbf{31.80}    \\
    \hline \hline
    \textbf{Mean (RMSE) } &  18.96    & \textbf{18.57}	& 30.88    & \textbf{30.32} \\
    \hline 
\end{tabular} \vspace{-3mm}
\end{table} \vspace{1mm}

\begin{figure}[htp]
    \centering
    \includegraphics[scale=.45]{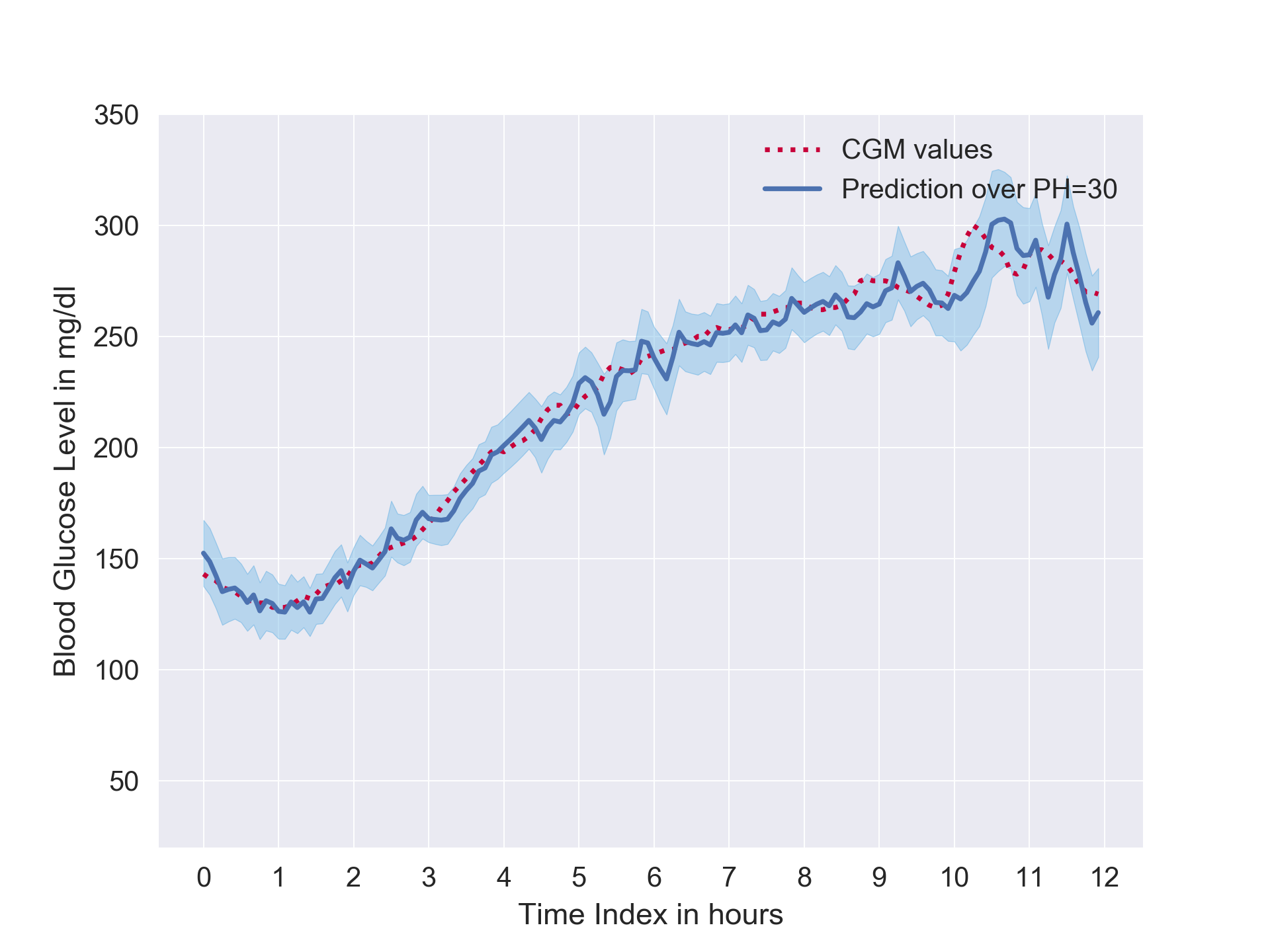}
    \hspace{1px}
    \includegraphics[scale=.45]{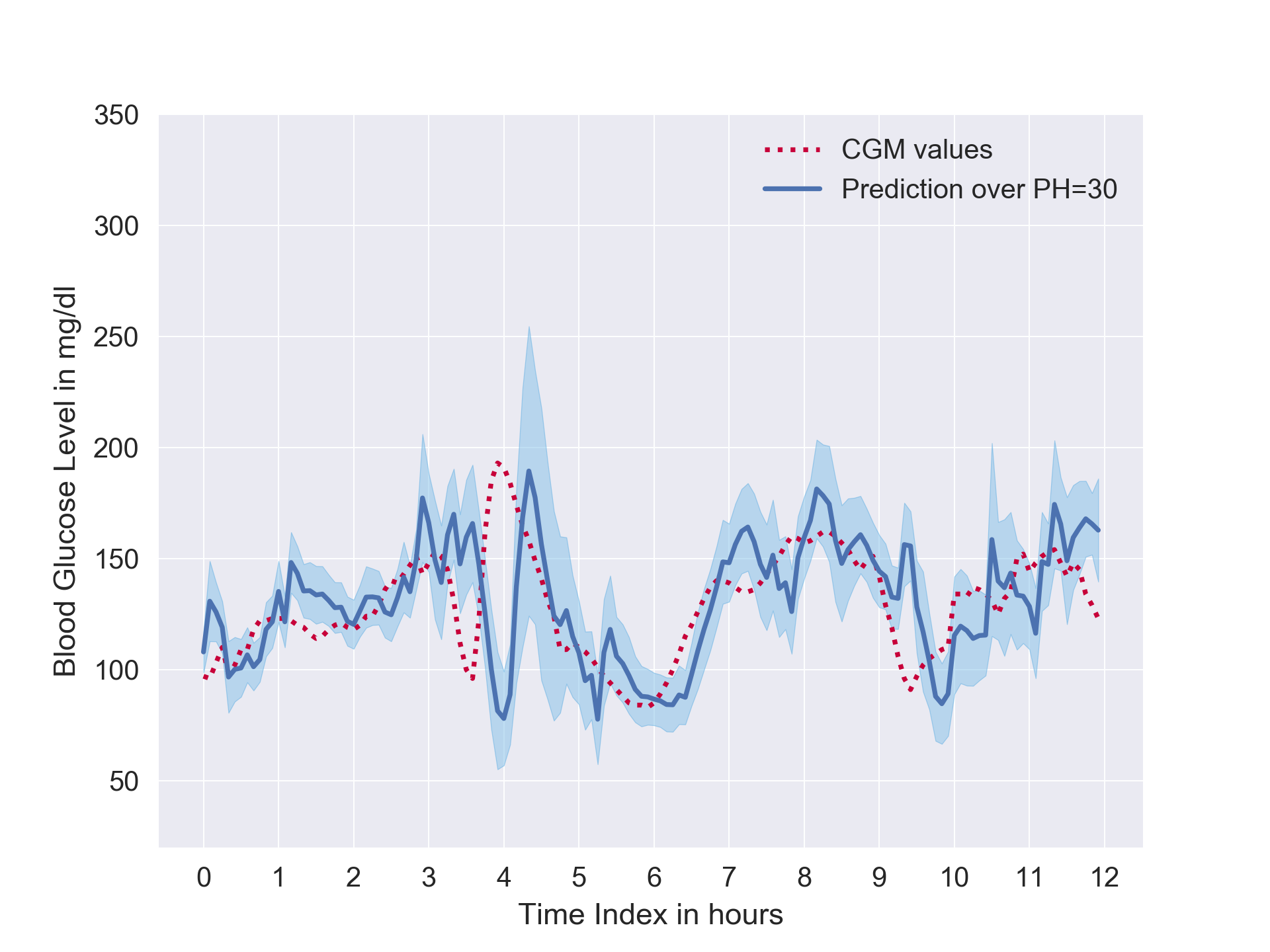}
    \caption{12-hours of prediction results over PH=30, for patient \#570 (top) and \#575 (bottom), are illustrated. Here red dots are CGM readings (unprocessed); those are ground truths. Where the blue line is the prediction curve, and the light blue region is the standard deviation.}
    \label{fig:prediction_Ph_30_stacked} \vspace{-4mm}
\end{figure}

The proposed model's (Stacked LSTM) predictive results for patient \#570 and \#575 over PH=30 minutes are illustrated in fig. \ref{fig:prediction_Ph_30_stacked}. The proposed model provides the lowest and the highest RMSE result for patient \#570 and patient \#575 respectively.



We also experiment with single GRU cell-based RNN for BG level prediction. However, both of the models with single and stacked LSTM cells provide better RMSE than the model with the GRU cell. The mean RMSE for the six patients' BG prediction are 20.07 and 31.12 for the PH=30 and 60 minutes respectively.   



\begin{figure*}[pos=htp,width=\textwidth,align=\centering]
    \centering
    \includegraphics[scale=.5]{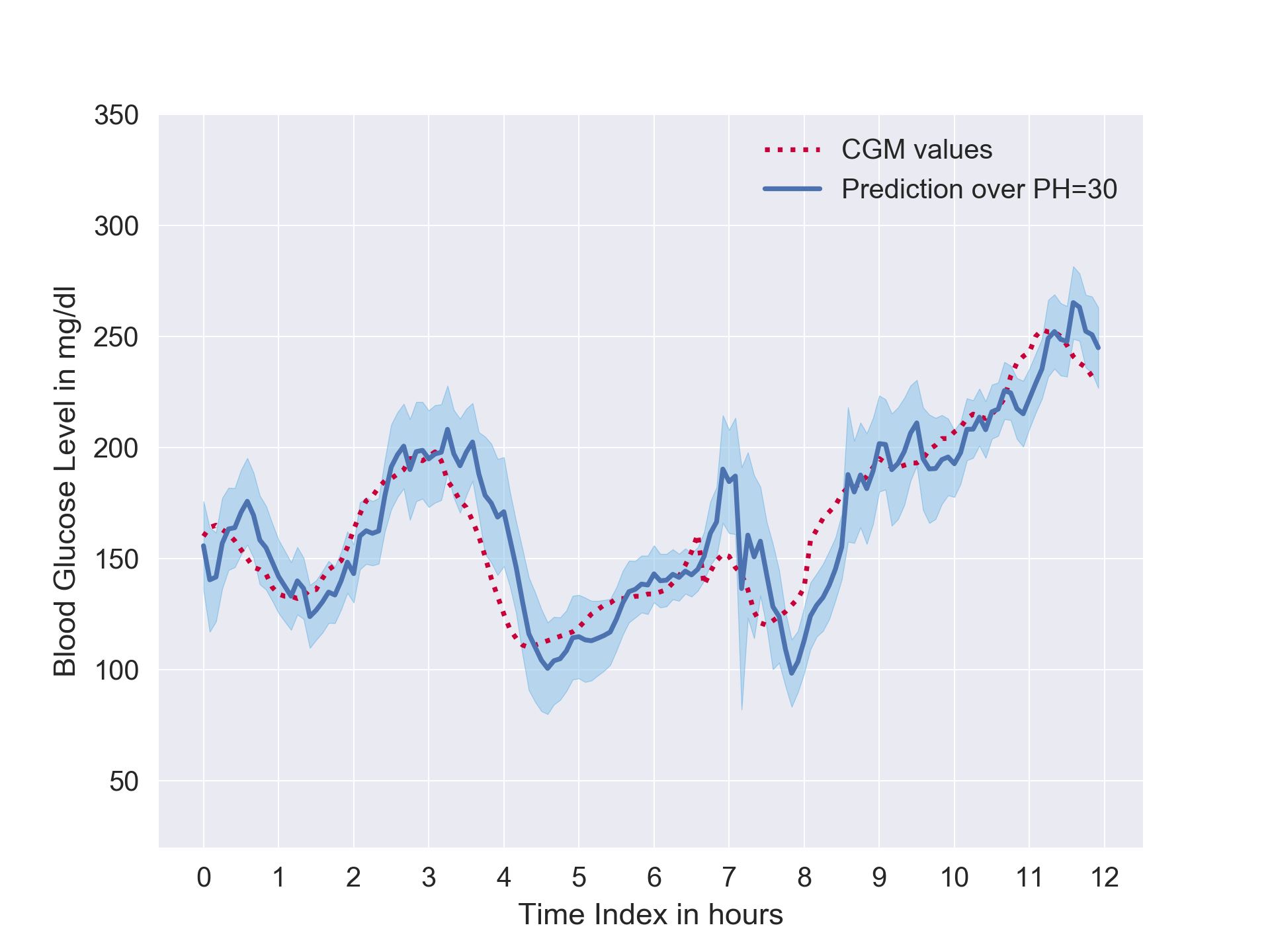}
    \hspace{1px}
    \includegraphics[scale=.5]{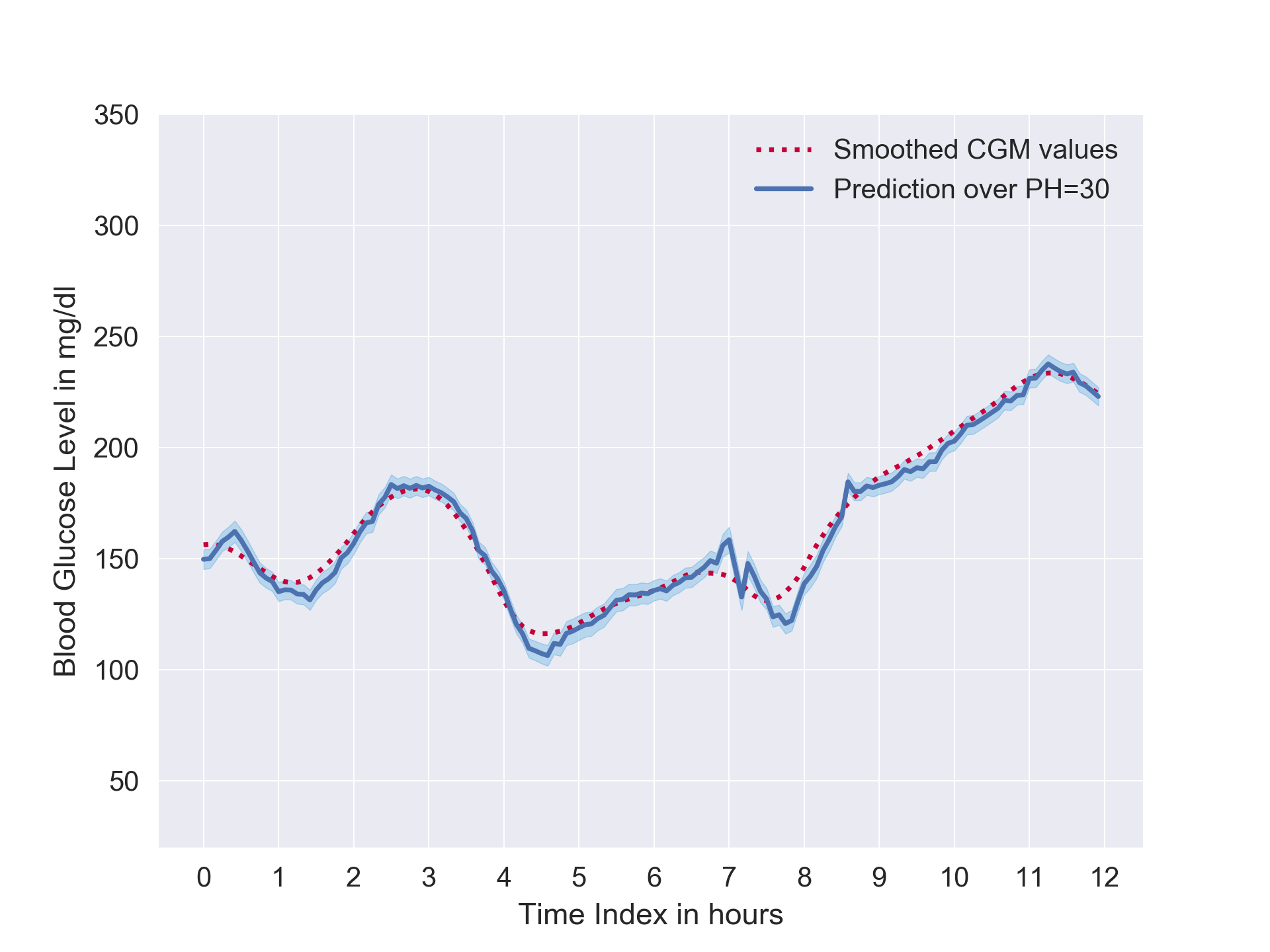}
    \caption{12-hour period prediction over PH=30, for patient \#563 with the model trained with unprocessed CGM readings (Left) and Kalman-smoothed CGM readings for sensor fault correction (Right) respectively.}
    \label{fig:Raw_Kalman_Prediction}
\end{figure*}

\subsection{Results for the dataset with Kalman Smoothed CGM readings}
In this section, the proposed model's predictive accuracy with stacked LSTM layers is evaluated for the preprocessed testing dataset with Kalman smoothing technique described in section \ref{kalman-smooting}. Note that, In our study, only CGM values are preprocessed, and the rest of the features remain the same. Then we estimate the RMSE of the model's prediction individually with preprocessed CGM values and raw CGM values, respectively, from the testing dataset as our goal is to lower the difference between the predicted CGM values and the real fingerstick blood glucose readings. The preprocessing techniques are described in Section \ref{data-processing}. Table \ref{table_RAW_VS_KALMAN_MODEL} presents the forecasting accuracy (RMSE) of the model trained with the preprocessed (CGM values) dataset using the Kalman smoothing technique and original (raw CGM values) dataset. 

\begin{table}[thp]
\renewcommand{\arraystretch}{1.3}
\caption{Accuracy (in terms of RMSE) of the final RNN model with stacked LSTM layers. The prediction result is for the dataset with raw CGM readings, and Kalman smoothed CGM values, respectively, over 30 minutes and 60 minutes of the prediction horizon.} 
\label{table_RAW_VS_KALMAN_MODEL} \vspace{-1mm}
\small
\centering
\begin{tabular}{ >{\centering\arraybackslash}m{1.4cm}  | >{\centering\arraybackslash}m{1.2cm} | >{\centering\arraybackslash}m{1.2cm} | >{\centering\arraybackslash}m{1.2cm} | >{\centering\arraybackslash}m{1.2cm} }
    \hline
    \multirow{3}{*}{\bf{Patient ID} } &
      \multicolumn{2}{c}{\bf{PH=30 min}} &
      \multicolumn{2}{c}{\bf{PH=60 min}} \\
    \cline{2-5}
    & \bf{Smoothed CGM} & \bf{Raw CGM} & \bf{Smoothed CGM} & \bf{Raw CGM}\\
    \hline \hline
     \# 559     & 4.73   &     17.85   &   16.17   &   31.55   \\
    \hline
     \# 563     & 5.74   &     18.65   &   16.31   &   30.42   \\
    \hline
     \# 570     & 4.81   &     15.94	&   14.22   &   25.74   \\
    \hline
     \# 575	    & 8.45   &     20.93	&   22.12   &   31.97   \\
    \hline
     \# 588	    & 5.10   &     17.71   &   15.73   &   30.45   \\
    \hline
     \# 591	    & 6.53   &     20.35	&   18.88   &   31.80    \\
    \hline \hline
    \textbf{Mean (RMSE)} &  \textbf{5.89}    & \textbf{18.57}	& \textbf{17.24}    & \textbf{30.32} \\
    \hline 
\end{tabular} \vspace{-4mm}
\end{table} \vspace{1mm}

Figure \ref{fig:Raw_Kalman_Prediction} demonstrates that preprocessing the CGM reading with Kalman smoothing, improves the prediction accuracy to a substantial extent.

\section{Analysis and Result Comparison}
From table \ref{table_raw_sigle_vs_stacked}, it is evident that, with a wider prediction horizon, the forecasting model becomes more complicated. However, we observe that deep approaches with stacked LSTM layers provide advantages over the shallow model with a single LSTM layer in forecasting BG level, particularly for a higher prediction horizon. As a result, we choose the deeper model for the final experiment.

From table \ref{table_RAW_VS_KALMAN_MODEL}, we can observe that prediction RMSE for each patient ranges from 15.94 to 20.94 and 4.73 to 8.54 for the model trained with raw CGM readings and processed CGM readings, respectively. The predictions for patients \#575 and \#591 are comparatively more inaccurate than other patients, whereas we achieved better RMSE for patient \#570 and \#588. There might be several reasons for the RMSE difference. The first reason is that \#575 and \#591 have a larger number of missing data. On the other hand, \#570 has the least missing data among all of the patients' test dataset. Another reason is that the fluctuation of CGM readings in the test dataset. It is noticeable that in the test dataset, \#575 and \#591 have substantially more fluctuation than \#570 and \#588. More specifically, the last portion of \#575 and the first portion of \#591 have more abrupt swings. Shown in Fig. \ref{fig:prediction_Ph_30_stacked}, 12-hours period BG level of \#570 contains less fluctuation than \#575. It is noteworthy that the prediction error is highest around the spikes and turning regions of the CGM trajectory. Furthermore, there is a marginal time delay in the prediction curve. This delay is responsible for the prediction error.

Such abrupt fluctuations in training and testing dataset make it difficult for the model to learn and predict the BG level accurately. One of the possible reasons behind such fluctuation is the complicated dynamic of the glucoregulatory system. Another possible reason is the sensor fault of the CGM system. We mitigate such undesirable errors by applying the Kalman smoothing technique on the dataset that makes CGM values less likely to have abrupt fluctuations. Subsequently, it boosts the learning capability of the model resulting in significantly better prediction accuracy. 

Fig. \ref{fig:Raw_Kalman_Prediction} illustrates how error correction of CGM reading, enhances the precision of the prediction curve for a particular 12-hours time window for patient \#563. Here we use patient \#563 as the dataset for this patient has average fluctuation. Note that, the SD (light blue region) is remarkably less for the prediction made with the model trained with processed data. Hence, the model trained with the error corrected dataset (Fig. \ref{fig:Raw_Kalman_Prediction} - Right), is capable of forecasting BG with more confidence than the model trained with the unprocessed dataset (Fig. \ref{fig:Raw_Kalman_Prediction} - Left). Moreover, we notice an improvement in the time delay in the proposed model.

\begin{table}[htbp]
\renewcommand{\arraystretch}{1.3}
\caption{Prediction accuracy comparison, in terms of RMSE, between the proposed model and models from related works for the OhioT1DM dataset (PH=30 minutes). Here, red and blue represents the best and second-best result, respectively.} 
\label{table_related_works} \vspace{-1mm}
\footnotesize
\centering
\begin{tabular}{ Z | Z | Z | Z | Z | Z }
    \hline
    \bf{Patient} & \bf{RMSE} &  \bf{RMSE} & \bf{RMSE of } & \bf{RMSE of } & \bf{RMSE of } \\
   & \bf{w/ KS} &  \bf{w/o KS} & \bf{ \cite{chen2018dilated}} & \bf{ \cite{martinsson2018automatic}} & \bf{ \cite{zhu2018deep}} \\
    
     \hline \hline
     \# 559     & \textcolor{Red}{\textbf{4.73}}   &     \textcolor{Blue}{\textbf{17.85}}   &   18.78   &   19.50  & 22.48   \\
     \hline
     \# 563     & \textcolor{Red}{\textbf{5.74}}   &     18.65   &   \textcolor{Blue}{\textbf{18.12}}   &   19.00  & 20.35   \\
    \hline
     \# 570     & \textcolor{Red}{\textbf{4.81}}   &     15.94	&   \textcolor{Blue}{\textbf{15.46}}  &   16.40  & 18.26   \\
    \hline
     \# 575	    & \textcolor{Red}{\textbf{8.45}}   &     \textcolor{Blue}{\textbf{20.93}}	&   22.83   &   24.80 & 25.65   \\
    \hline
     \# 588	    & \textcolor{Red}{\textbf{5.10}}   &     \textcolor{Blue}{\textbf{17.71}}    &   17.72   &   19.30  & 21.69  \\
    \hline
     \# 591	    & \textcolor{Red}{\textbf{6.53}}   &     \textcolor{Blue}{\textbf{20.35}}	&   21.34   &   25.40  & 24.59  \\
    \hline \hline
    \textbf{Mean} &  \textcolor{Red}{\textbf{5.89}}    & \textcolor{Blue}{\textbf{18.57}}	& 19.04    & 20.73  & 22.17 \\
    \hline
\end{tabular} \vspace{-4mm}
\end{table} \vspace{1mm}

We provide a performance comparison of our work with related works for the OhioT1DM dataset in table \ref{table_related_works}. For comparison, we consider previous works \cite{chen2018dilated}, \cite{martinsson2018automatic}, \cite{zhu2018deep} those uses the OhioT1DM dataset for result evaluation. Table \ref{table_related_works} demonstrates that our proposed model with KS provides the best accuracy for every patient among all other related works. Even without utilizing KS, we achieved the topmost accuracy for the four patients out of six. For patient \#563 and \#570, the work \cite{chen2018dilated} has slightly better accuracy than ours. \cite{martinsson2018automatic} and \cite{chen2018dilated} use LSTM and dilated RNN for BG prediction respectively whereas \cite{zhu2018deep} uses a convolutional neural network (CNN) trained with the OhioT1DM dataset for the BG level prediction. To the best of our knowledge, we achieved the leading average prediction accuracy for the OhioT1DM dataset.


\section{Conclusion}
This work investigated methods for more accurate blood glucose prediction. We demonstrated that preprocessing the CGM readings with Kalman smoothing for sensor error correction could be useful for improving the robustness of the BG prediction. We utilized different physiological information such as meal, insulin, aggregations of step count, and preprocessed CGM data in our method. We proposed a novel approach leveraging the stacked LSTM based deep RNN model to improve the BG prediction accuracy in this paper. It is evident from our study that preprocessing the CGM values with Kaman smoothing makes the BG prediction curve less uncertain and less fluctuating. Our proposed approach provides more reliable predictions than traditional methods while we assumed fingerstick BG readings as the ground truth in our experiment. We want to lower the difference between the predicted CGM values and the real fingerstick blood glucose readings. The BG prediction with Kalman smoothed CGM data is closer to the actual BG level (The Fingerstick BG reading) than without the Kalman filter. This more accurate prediction can aid diabetes patients to avoid adverse glycemic events. Our proposed methodologies could be employed to get insight into T1D patient's future BG level trends that might result in a more dependable diabetes management system.

\section*{Acknowledgment}
This work is supported in part by US NSF under grants CNS-1812553, OIA-1946231.

\printcredits

\bibliographystyle{cas-model2-names}

\bibliography{main.bib}

\vskip3pt




\end{document}